\documentclass[10pt,twocolumn,letterpaper]{article}

\usepackage{iccv}
\usepackage{times}
\usepackage{epsfig}
\usepackage{graphicx}
\usepackage{amsmath}
\usepackage{amssymb}
\usepackage{multirow}
\usepackage{graphicx}
\usepackage[euler]{textgreek}
\usepackage{algorithm}
\usepackage{algpseudocode}
\usepackage{booktabs}

\usepackage[breaklinks=true,bookmarks=false]{hyperref}

\usepackage{pifont}
\newcommand{\cmark}{\ding{51}}%
\newcommand{\xmark}{\text{\ding{55}}}

\newcommand{\figref}[1]{Fig.~\ref{#1}}
\newcommand{\figureref}[1]{Figure~\ref{#1}}
\newcommand{\tabref}[1]{Tab~\ref{#1}}
\newcommand{\tableref}[1]{Table~\ref{#1}}

\iccvfinalcopy
\begin{document}

\title{Implicit Stacked Autoregressive Model for Video Prediction}

\author{Minseok Seo\hspace{4mm} Hakjin Lee\hspace{4mm} Doyi Kim\hspace{4mm} Junghoon Seo\\
SI Analytics\\
70, Yuseong-daero 1689beon-gil, Yuseong-gu,
Daejeon, Republic of Korea\\
{\tt\small \{minseok.seo, hakjinlee, doyikim, jhseo\}@si-analytics.ai}
}

\maketitle

\begin{abstract}
Future frame prediction has been approached through two primary methods: autoregressive and non-autoregressive.
Autoregressive methods rely on the Markov assumption and can achieve high accuracy in the early stages of prediction when errors are not yet accumulated.
However, their performance tends to decline as the number of time steps increases.
In contrast, non-autoregressive methods can achieve relatively high performance but lack correlation between predictions for each time step.
In this paper, we propose an Implicit Stacked Autoregressive Model for Video Prediction (IAM4VP), which is an implicit video prediction model that applies a stacked autoregressive method.
Like non-autoregressive methods, stacked autoregressive methods use the same observed frame to estimate all future frames.
However, they use their own predictions as input, similar to autoregressive methods.
As the number of time steps increases, predictions are sequentially stacked in the queue.
To evaluate the effectiveness of IAM4VP, we conducted experiments on three common future frame prediction benchmark datasets and weather\&climate prediction benchmark datasets.
The results demonstrate that our proposed model achieves state-of-the-art performance.
\end{abstract}

\section{Introduction}
Video prediction involves the task of predicting a sequence of future frames based on past input frames, with the aim of modeling spatio-temporal representation learning.
The common deep learning architecture for this task includes an \textit{Encoder} that extracts representations from a single static image, and a \textit{Decoder} that transforms these representations into the required output~\cite{ronneberger2015u}.
However, video prediction tasks require an additional module, referred to as the \textit{Predictor} in this paper, which is also known as the \textit{Translator} and is responsible for learning spatio-temporal information between consecutive images.
Depending on the approach used to structure the \textit{Predictor} module, past studies can be classified into two types: autoregressive and non-autoregressive models.
Autoregressive models rely on predicted results from past frames to determine the current model output, whereas non-autoregressive models do not have this dependence.
\begin{figure}[t!]
    \centering
    \includegraphics[width=1.0\columnwidth]{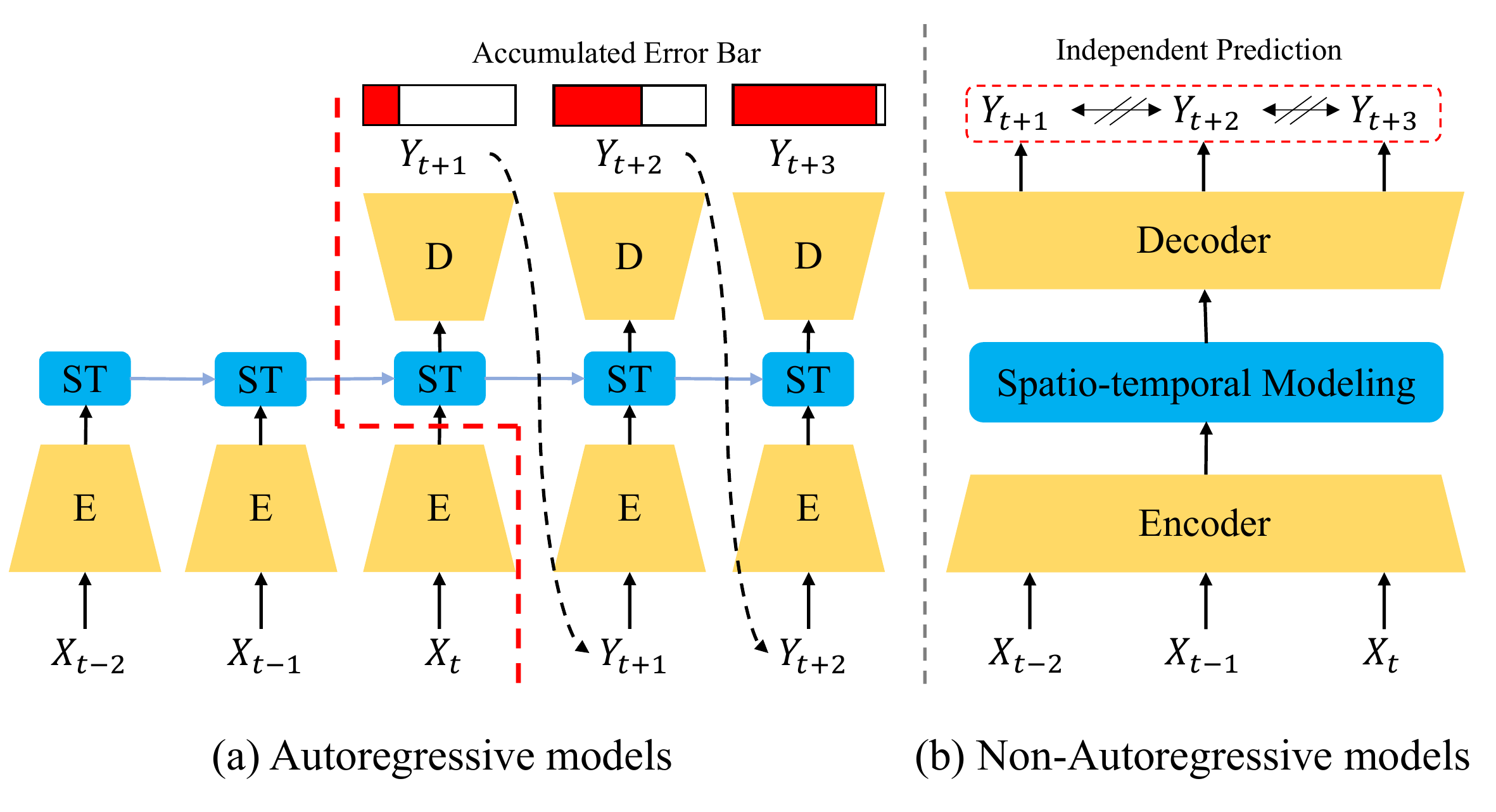}
    \caption{An overview of autoregressive and non-autoregressive models. Autoregressive models have the limitation of accumulating errors, while non-autoregressive models have a limitation in that each output is independent of the context of previous time steps.}
    \label{fig:motivation}
\end{figure}

Autoregressive models, also known as Single-In-Single-Out (SISO) architectures, such as ConvLSTM~\cite{shi2015convolutional} and PredRNN~\cite{wang2017predrnn}, aim to model spatio-temporal memory states by passing the recursively previous frames, regardless of whether they come from actual inputs or predictions for the previous frames.
However, errors can accumulate as they recursively use their prediction results as inputs when predicting future frames, leading to decreased accuracy over time.
This issue is illustrated in ~\figref{fig:motivation}-(a) and ~\tableref{table:properties}.

Recent studies, such as SimVP~\cite{gao2022simvp} and MIMO-VP~\cite{ning2023mimo}, have focused on non-autoregressive models, which were previously referred to as Multiple-In-Multiple-Output (MIMO) architectures. These models have demonstrated remarkable performance gains without the need for complex recursive architectures. They encode a spatio-temporal representation by stacking feature maps of all input frames in the \textit{Translator} module and decode it to multiple future frames simultaneously. Specifically, Ning \textit{et al.}~\cite{ning2023mimo} highlighted the error accumulation problem that arises in Single-In-Single-Out (SISO) models and emphasized the MIMO architecture that predicts future frames in a one-shot fashion, thereby eliminating this problem. It should be noted, however, that MIMO models do not follow the Markov assumption, as they do not capture the correlation between each prediction and are independent, as illustrated in ~\figref{fig:motivation}-(b).

We conducted a comparative experiment between MIMO and MISO architectures to evaluate the effectiveness of MIMO in addressing the error accumulation problem, as described in Section~\ref{sec:miso-power}.
The experimental results show that while MIMO architectures provide advantages in the baseline model, they are not necessarily critical for mitigating error accumulation.
Notably, the results demonstrate that the MISO-Multi Model, a variant of the MISO architecture, outperforms the MIMO architecture, suggesting that alternative approaches can achieve superior results in addressing error accumulation.
Based on these findings, we propose an implicit video prediction model that mimics the MISO-Multi Model and leverages the versatility of autoregressive models.
Additionally, we introduce a stacked autoregressive method to address the lack of context prediction and output correlation in non-autoregressive models.
The proposed Implicit Stacked Autoregressive Model for Video Prediction (IAM4VP), which incorporates these two novel methods, achieves state-of-the-art performance on three widely used future frame prediction benchmark datasets and two benchmark datasets for weather and climate prediction.

\begin{table}[tp]
  \centering
  \scalebox{0.68}{
  \begin{tabular}{l|c|c|c}
    \toprule
    \textbf{Properties} & \multicolumn{1}{c|}{\begin{tabular}[c]{@{}c@{}}\textbf{Existing}\\ \textbf{Autoregressive}\end{tabular}} & \multicolumn{1}{c|}{\begin{tabular}[c]{@{}c@{}}\textbf{Existing}\\ \textbf{Non-autoregressive}\end{tabular}} & \begin{tabular}[c]{@{}c@{}}\textbf{IAM4VP}\\\textbf{(Ours)}\end{tabular} \\ \midrule
    No Error Accumulation       & \xmark & \cmark & \cmark \\ 
    Correlation btw Predictions & \cmark & \xmark & \cmark \\
    \midrule
    Number of Input&Single&Multi&Multi \\
    Number of Output&Single&Multi&Single \\
    \bottomrule
  \end{tabular}
  }
  \caption{Properties of current video prediction models.}
  \label{table:properties}
\end{table}

Our contributions can be summarized as follows:

\begin{itemize}
\setlength\itemsep{0em}
\item We propose a novel autoregressive model that predicts future frames recursively with a Multiple-In-Single-Out design to avoid the error accumulation problem.
\item We propose a way to emulate multiple models using a temporal embedding and a spatio-temporal predictor.
\item We demonstrate that our Implicit Stacked Autoregressive Model for Video Prediction (IAM4VP) achieves state-of-the-art performance on three widely-used video prediction benchmark datasets and two benchmark datasets for weather and climate prediction.
\end{itemize}
\begin{table*}[h!]
{
\begin{center}
\renewcommand{\arraystretch}{1.2}
\resizebox{1.6\columnwidth}{!}{%
\begin{tabular}{l|cc|cccccccccc|c}
\hline \hline
\multicolumn{1}{c|}{\multirow{2}{*}{\textbf{Method}}} & \multirow{2}{*}{\textbf{\#Param. (M)}} & \multirow{2}{*}{\textbf{Training Epoch}} & \multicolumn{10}{c|}{\textbf{Time Step}} & \multirow{2}{*}{\textbf{MSE}} \\ \cline{4-13}
\multicolumn{1}{c|}{}                                 &                                        &                                          & 1  & 2  & 3 & 4 & 5 & 6 & 7 & 8 & 9 & 10 &                               \\ \hline
SimVP-S (MIMO)                                        &20.4                                       & 2K                                         & 11.6   & 14.7   & 17.5  & 19.6  & 22.0  & 24.4  &  27.1 & 29.4  & 32.6  &  36.5  &  23.5                             \\
SimVP-S (MIMO)*10                                     & 20.4                                       & 20K                                         & 8.8    & 11.9   & 15.6  & 17.3  & 19.2  & 21.4  & 25.7  & 28.8   & 30.3  & 35.2   & 21.4                              \\
SimVP-L (MIMO)                                        &   53.5                                     &  2K                                        &  13.7  & 17.5   & 18.1  & 20.2 &  24.1 &  26.6 &  29.8 &  33.1 &  36.5 & 38.2   &   25.7                            \\ \hline
SimVP-S (MISO-Multi Model)                            & 20.4*10                                       &      20K                                    & \textbf{8.3}   &  \textbf{10.9}  & \textbf{13.1}  & \textbf{15.6}  &   \textbf{17.8}&  \textbf{20.0} & \textbf{22.4}  &  \textbf{24.5} & \textbf{26.1}  & \textbf{28.7}   &     \textbf{18.7}                          \\
SimVP-S (MISO-Autoregressive)                         &  20.4                                      &     2K                                     &  8.3   & 13.4   &  19.2 &  24.5 & 30.3  &36.2   & 42.6   & 48.7   &  55.2 &  62.3  & 34.1                                \\ \hline \hline
\end{tabular}%
}
\end{center}
\caption{The result of comparing the efficiency of each method after changing the previous architecture SimVP model to Multiple-In-Single-Out Multi Model (MISO-Multi Model) and Multiple-In-Single-Out-autoregressive (MISO-Autoregressive) structure. All experiments were conducted on the Moving MNIST dataset.
}
\label{tab:motivation}
}
\end{table*}
\section{Related Work}
Estimating a future video from past video sequences can be divided into two sub-problems: future frame prediction and future frame generation.
Future frame prediction \cite{shi2015convolutional, wang2018predrnn++, gao2022simvp} involves estimating a future sequence with the highest probability of being observed given a past video sequence, while future frame generation \cite{xu2018video, denton2018stochastic, babaeizadeh2018stochastic, franceschi2020stochastic, akan2021slamp, voleti2022mcvd} involves training and sampling the distribution of future video frames based on given past video sequences.
In this work, we focus primarily on future frame prediction, which we refer to as video prediction for brevity.

Video prediction can be achieved using either an autoregressive model or a non-autoregressive model, depending on whether the model uses the result of previous estimations as input to predict the future frame. Our work focuses on a type of autoregressive models that have been inspired by recent state-of-the-art non-autoregressive models.

\subsection{Autoregressive Video Prediction Models} 
Autoregressive video prediction \cite{shi2015convolutional, lu2017flexible, yu2019crevnet, wang2019eidetic, wang2019memory, castrejon2019improved, su2020convolutional, guen2020disentangling, wu2021motionrnn, chang2021mau, wang2022predrnn, chang2022strpm} is a common approach in video prediction research, where models are used to predict future frames based on the current frame. One advantage of this method is that it can naturally correspond to long-term predictions for an infinite time horizon. Since inference is processed sequentially, the memory requirement does not increase with the length of the input, which is particularly useful for dealing with long input and output sequences. The challenge, however, is to retain the relevant historical information from past sequences, which has led to research focus on storing and updating this historical information state. Although various models like ConvLSTM \cite{shi2015convolutional}, PredRNN \cite{wang2022predrnn, wang2018predrnn++}, and MIM \cite{wang2019memory} have attempted to capture both spatial and temporal representations, they still struggle to extrapolate longer sequences and capture long-term dependencies.

One of the primary causes of the autoregressive model's inability to capture longer sequences is the error accumulation problem. The model's error propagates at each time step, leading to out-of-distribution predictions and performance degradation. Recent attempts to solve this problem, called the hierarchical approach, involve learning high-level semantic structures such as human pose, segmentation map, and relational layout. This semantic information is used as a condition for predicting future frames \cite{villegas2017learning, lee2021revisiting, bodla2021hierarchical}, or as an auxiliary for learning a good representation in the training phase \cite{villar2022MSPred}. However, these models require additional semantic label information and still suffer from error accumulation over time.


\subsection{Non-autoregressive Video Prediction Models} 
Non-autoregressive video prediction models \cite{liu2017video, aigner2018futuregan, gao2022simvp, xu2018predcnn, gao2022earthformer, ning2023mimo} have received less attention in the literature compared to their autoregressive counterparts. However, recent studies have highlighted their potential for achieving high performance with simplicity. Most notably, non-autoregressive models do not suffer from the error accumulation problem associated with autoregressive models. For example, Deep Voxel Flow \cite{liu2017video} and FutureGAN \cite{aigner2018futuregan} are non-autoregressive models that use 3D convolution autoencoders and spatio-temporal 3D convolutions, respectively. Another non-autoregressive model, called PredCNN \cite{xu2018predcnn}, incorporates a cascade multiplicative unit to capture temporal dependencies using a CNN-based architecture.

Recently, Gao et al. \cite{gao2022simvp} proposed SimVP, a non-autoregressive model based entirely on CNNs that uses a Multiple-In-Multiple-Out (MIMO) predictive strategy to capture temporal relationships across frames. SimVP achieves significant performance using only basic CNNs. Ning et al. \cite{ning2023mimo} aimed to develop a high-performing MIMO model that fully exploits the strength of MIMO architectures. To achieve this, they proposed a new MIMO architecture, MIMO-VP, which extends the pure Transformer with local spatio-temporal blocks and a new multi-output decoder. MIMO-VP outperforms state-of-the-art models and achieves remarkable performance on four benchmarks.

Despite its advantage in prediction performance, non-autoregressive models have a disadvantage in that memory requirements increase rapidly as the length of the input and output sequences increase. Moreover, unlike autoregressive models, they generally do not model temporal correlations between predictions. This paper reexamines important performance factors that have been claimed (but not yet verified) in studies of state-of-the-art non-autoregressive video prediction models such as SimVP and MIMO-VP.



\begin{figure*}[t!]
    \centering
    \includegraphics[width=1.5\columnwidth]{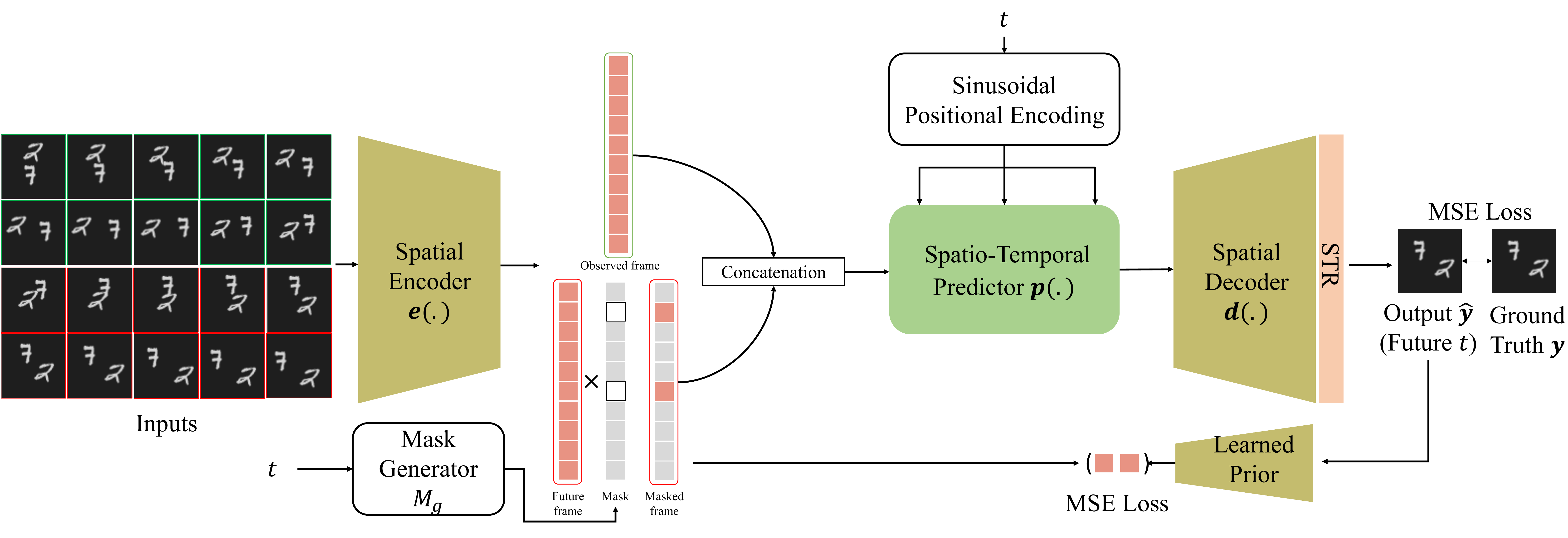}
    \caption{An overview of IAM4VP's training process. IAM4VP is an implicit model for time step $t$. IAM4VP consists of a spatial encoder $e(\cdot)$, a spatio-temporal predictor $p(\cdot)$, a spatial decoder $d(\cdot)$, and a spatial temporal refinement module (STR). During the IAM4VP training process, a prior $M_{g}$ and sinusoidal positional encoding values for time step $t$ are generated and input to the model. Finally, an additional encoder receives $\hat{y}$ and is fine-tuned to minimize the mean squared error (MSE) loss between the features of the ground truths $y$.}
    \label{fig:divmodnet}
\end{figure*}

\section{Exploring the Potential of MISO}
\subsection{Analysis}
\label{sec:miso-power}

Recent studies suggest that non-autoregressive methods, specifically the MIMO model, may provide an advantage in video prediction problems. Our primary research questions are whether a MIMO architecture is essential for achieving high performance in video prediction and whether the multiple-output component is critical in the MIMO architecture. To answer these questions, we compared autoregressive and non-autoregressive models with the same structure and hyperparameter settings. We modified the SimVP MIMO model to create the MISO-autoregressive and MISO-Multi Model and conducted ablation experiments to confirm their efficacy.

Note that the MISO-autoregressive model predicts the future frame $t_{n+1}$ by inputting time step points $B={ t_{0}, t_{1}, ..., t_{n}}$ and removing them from the queue in a first-in-first-out (FIFO) manner in the next step. In addition, the MISO-Multi Model also uses multiple inputs like the MIMO model but models ${ f_{n+1}(B), f_{n+2}(B), ..., f_{n+m}(B)}$, where $m$ is the number of target future frames that are specifically configured.

Table~\ref{tab:motivation} presents the effectiveness comparison of the MIMO, MISO-autoregressive, and MISO-Multi Model. Our experiments reveal that Mean Squared Error (MSE) decreases as time step increases in all experiments. Moreover, the Multi Model exhibits the highest performance. These experimental results demonstrate that the multi-input components, rather than the multi-output, significantly influence performance in the MIMO method.

Furthermore, the autoregressive model exhibits the lowest performance, indicating that the error accumulation problem is critical in future frame prediction tasks. Hence, when designing a model for future frame prediction, it is essential to consider the design of a multi-input structure that does not accumulate errors.

\subsection{Motivation}

The initial experiments in Section~\ref{sec:miso-power} suggest that the MISO model has the potential for high performance in video prediction problems.
However, the MISO-Multi Model is computationally inefficient since it requires learning and inference of multiple models.
Additionally, the MISO-Multi Model lacks modeling of time dependence since it does not use previous or subsequent predicted frames when inferring each future timestamp.
In contrast, the MISO-Autoregressive model models time dependence but suffers from long-term error accumulation, resulting in performance degradation.

Despite these drawbacks, the MISO model shows the most potential for video prediction, and this study aims to leverage it.
Therefore, our novel model for the video prediction problem is designed based on the following principles: First, we use a Multiple-In-Single-Out model.
Second, for computational efficiency, we use a single model instead of multiple models. 
Third, we adopt an autoregressive approach to model spatio-temporal dependence in prediction.
Finally, our model should demonstrate state-of-the-art performance compared to existing SISO and MIMO models.
In summary, our goal is to build an \textbf{autoregressive video prediction model} based on a \textbf{Multiple-In-Single-Out architecture} that achieves \textbf{state-of-the-art performance}.

\section{Method}
\label{sec:method}
This section presents the Implicit Stacked Autoregressive Model for Video Prediction (IAM4VP), illustrated in ~\figref{fig:divmodnet}.
IAM4VP comprises multiple modules, including an encoder, observed and future queues, predictor, decoder, additional encoder, and a spatio-temporal refinement module called STR~\cite{seo2022simple}.
Given an input $x$ and target time step $t$, IAM4VP generates a prediction for $y$ by sequentially passing $x$ through the encoder-predictor-decoder.
The resulting prediction $\hat{y}$ is refined through STR and then mapped to the ground truth $y$ feature using an additional encoder before being stacked in the future queue.

\subsection{Baseline}
\paragraph{Problem Definition}
Given a dataset $\mathcal{D} ={(x_{i}, y_{i})}_{i=1}^{N}$, the goal of the future frame prediction model $F(\cdot)$ is to map the input $x \in \mathbb{R}^{C \times T \times H \times W }$ to the target output $y \in \mathbb{R}^{C \times \hat{T} \times H \times W}$. The learnable parameters of the model, denoted by $\textTheta$, are optimized to minimize the following loss function:
\begin{equation}
\textTheta = \min_{\textTheta}\frac{1}{N}\underset{x, y \in D}{\sum}L(F_{\textTheta}(x),y).
\label{eq:object}
\end{equation}
Here, $x$ is the observed frame, $y$ is the future frame, $C$ is the number of channels, $T$ is the observed frame length, $\hat{T}$ is the future frame length, $H$ is the height, and $W$ is the width of the frames.
Different loss functions $L$ can be used for optimization, such as mean squared error (MSE)~\cite{gao2022simvp}, mean absolute error (MAE)~\cite{ning2023mimo}, smooth loss~\cite{seo2022domain}, or perceptual loss~\cite{shouno2020photo}.
In this work, we use only MSE loss, which is the most commonly used loss function for future frame prediction tasks.
\paragraph{Encoder-Predictor-Decoder}
In the task of future frame prediction, the current leading architecture design is the \textit{encoder-predictor-decoder} structure~\cite{gao2022simvp}.
The encoder extracts features from the observed frame $x$ and passes them to the predictor, which generates future features.
Finally, the decoder decodes the future features to reconstruct the future frame $y$.
IAM4VP also adopts the encoder-predictor-decoder structure, which consists of {\fontfamily{qcr}\selectfont(Conv, LayerNorm, SiRU)} for the encoder, {\fontfamily{qcr}\selectfont(Conv, LayerNorm, SiRU, PixelShuffle)} for the decoder, and ConvNeXt blocks for the predictor.
\subsection{Implicit MISO Architecture}
Through the experimental results shown in \tabref{tab:motivation}, we found that the \textit{MISO-Multi Model} is the best performing strategy among those evaluated for predicting future frames.
However, the \textit{MISO-Multi Model} strategy is inefficient because it requires training multiple models, one for each time step in the target sequence $\hat{T}$, and the total number of model parameters increases with the length of $\hat{T}$.
Furthermore, the MISO-Multi Model does not account for the correlation between predictions in modeling just like MIMO models.
To address these issues, we propose the implicit MISO architecture, inspired by the implicit neural network~\cite{sitzmann2020implicit}.

The implicit MISO model $F(\cdot)$ is trained to predict the future frame $y_{t}$ by taking as input the observed frame $x$ and the target time step $t \in \hat{T}$.
Thus, given a pair of input-output $(x, y)$, the model $F(\cdot)$ is trained to minimize the following loss function:
\begin{equation}
  \textTheta = \min_{\textTheta}\frac{1}{N}\underset{x, y \in D}{\sum}\underset{t \in \hat{T}}{\sum}L(d_{\textTheta}(p_{\textTheta}(e_{\textTheta}(x), s(t))),y_{t}),
  \label{eq:object}
\end{equation}
where $e(\cdot)$ is the encoder, $p(\cdot)$ is the predictor, $d(\cdot)$ is the decoder, and $s(\cdot)$ is the sinusoidal positional encoding.
The target time step $t$, which has been positional encoded through $s(\cdot)$, is added to each layer of $p(\cdot)$ after the features are extracted through an MLP layer composed of two fully connected (FC) layers and one GELU activation \cite{hendrycks2016gaussian}.

When training IAM4VP, we use a validation error-based sampling method for each time step $t$ during each epoch, to prevent $t$ from being sampled in a specific interval or inefficiently sampled.
The sampling strategy for $t$ follows the formula below:
\begin{equation}
  B = \{(\frac{(MSE_{i})^{\frac{1}{\alpha}}}{\sum_{i=0}^{\mathbf{\hat{T}}}({MSE_{i}})^{\frac{1}{\alpha}}}\}_{i=1}^{\hat{T}}
\label{eq:sample}
\end{equation}
where $B$ is the timestep sampling probability, $\alpha$ is the sharpening coefficient. We set $\alpha$ to 1 in all experiments.
\subsection{Stacked Autoregressive Strategy}
Non-autoregressive methods such as implicit MISO and MIMO achieve high accuracy because errors do not accumulate, unlike autoregressive methods. However, since their outputs are independent, they may generate outputs that are unrelated to the surrounding context at a specific time step $t$.

To combine the best parts of both approaches, we propose a stacked autoregressive strategy. The strategy is divided into an observed frame queue and a future frame queue. The observed frame queue remains unchanged as the time step increases and always has the same encoding value. The future frame queue is initially filled with all zero tensors. The predicted values are then gradually filled in the future frame queue as the time step increases autoregressively. Finally, the features accumulated in the observed queue and the future frame queue (feature $\hat{Y}$ corresponding to time steps $t_{n+1}$ to $t_{n+m-k}$) are concatenated and fed into the predictor $p(\cdot)$. The output of the predictor $p(\cdot)$ is then used with the decoder $d(\cdot)$ to predict $y_{t}$.

\begin{figure}[t!]
    \centering
    \includegraphics[width=0.8\columnwidth]{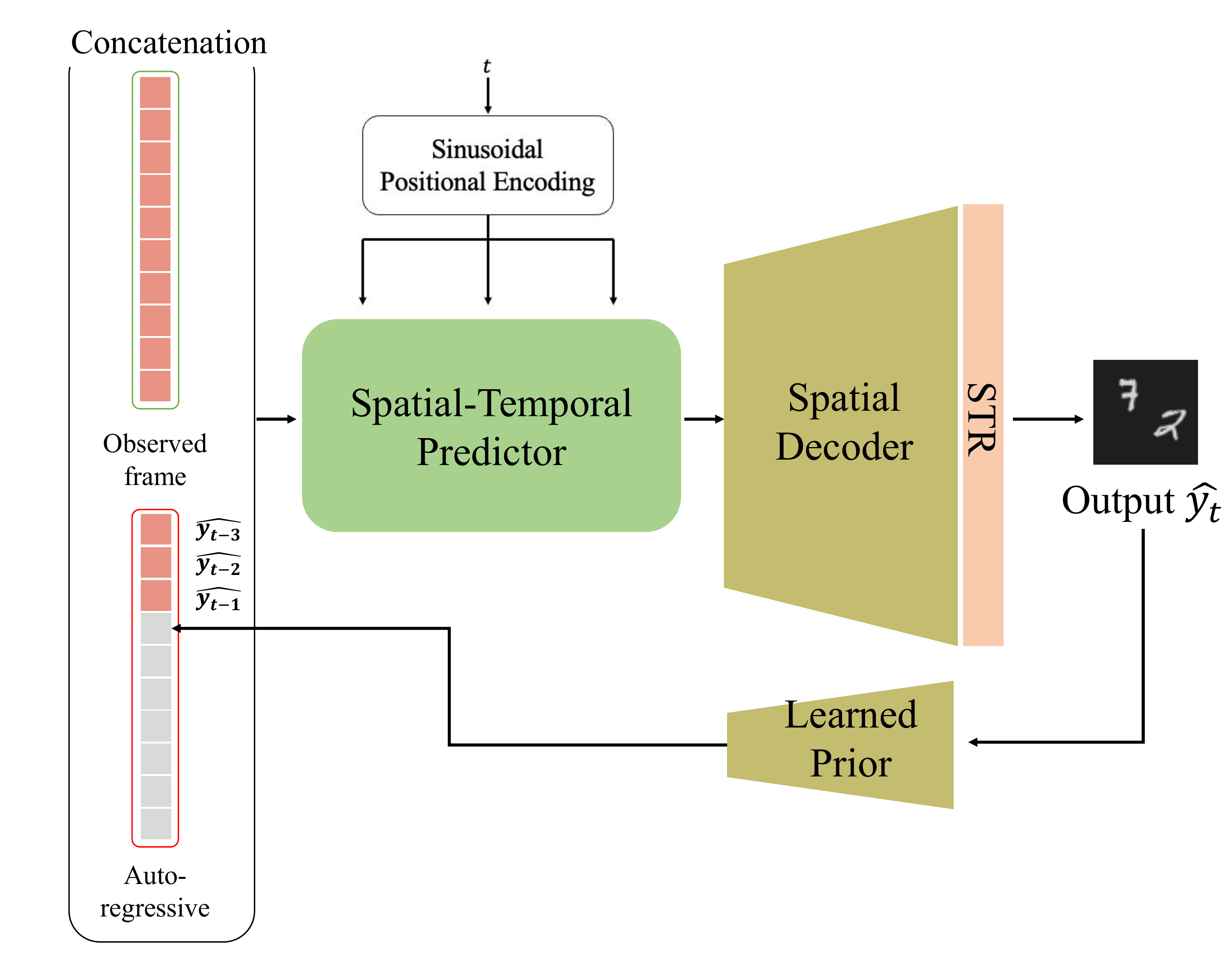}
    \caption{The overview of IAM4VP inference process.}
    \label{fig:infer}
\end{figure}

\begin{algorithm}
    \caption{Training Strategy}\label{euclid}
    \begin{algorithmic}
        \State \textbf{Input:} Observed frame $x$, future frame $y$, target time step $\hat{T}$, encoder $e(\cdot)$, predictor $p(\cdot)$ and decoder $d(\cdot)$, observed queue $Q_{o}$ and future queue 
        \For {$x_{i}, y_{i}$ in $\mathcal{B}$}
        \State $t = random(\hat{T})$ \Comment{$t \in \hat{T}$}      
        \State $Q_{o}, Q_{f} = e(x_{i}), e(y_{i})$ \Comment{queue generation}
        \State $Mask = M_{g}(t)$ \Comment{Mask generation according to $t$}
        \State $Q_{f} = Mask \times Q_{f}$ \Comment{masked $Q_{f}$}
        \State $\hat{y}_{t} = d(p(Concat(Q_{o},Q_{f}), t))$ \Comment{prediction}
        \State $loss = L(\hat{y}_{t}, y_{t})$
        \EndFor
    \end{algorithmic}
    \label{alg:sampling}
\end{algorithm}

To reduce the computational cost of training autoregressive methods, we employ the ground truth future frames as the future queue during the training stage. Instead of using feature vectors extracted from predictions on time steps from $t_{n+1}$ to $t_{n+m-k}$, we extract feature vectors directly from the future ground truths at time steps from $t_{n+1}$ to $t_{n+m}$. Additionally, we introduce a technique inspired by~\cite{voleti2022mcvd} that randomly masks some parts of the future queue with zero tensors during the training phase. This approach enhances the model's ability to predict the target time step specified in the sinusoidal positional embedding even when incomplete context information is provided from the future queue. Algorithm~\ref{alg:sampling} presents our overall training algorithm for IAM4VP, including the implicit MISO and stacked autoregressive strategies.
The mask generator $M_{g}$ follows the rule according to $t$:
\begin{equation}
    Mask = \begin{cases}
        Random(t) & t > Index(Q_{f}), \\
        0, & t \leq Index(Q_{f}),  
    \end{cases}
\label{eq:mask}
\end{equation}
where $Random(\cdot)$ is a function that selects a random number of random indices.

\begin{table*}[t!]
\begin{center}
\resizebox{1.5\columnwidth}{!}{%
\begin{tabular}{l|ccc|ccc|ccc}
\hline
\hline
            & \multicolumn{3}{c|}{\textbf{Moving MNIST}} & \multicolumn{3}{c|}{\textbf{TrafficBJ}} & \multicolumn{3}{c}{\textbf{Human 3.6}} \\ \hline
\textbf{Method}      & MSE       & MAE       & SSIM      & MSE $\times 100$      & MAE      & SSIM     & MSE$ / 10$      & MAE $ / 100$    & SSIM     \\ \cline{1-10} 
ConvLSTM~\cite{shi2015convolutional} &   103.3        &  182.9         &   0.707        &  48.5        &   17.7       &   0.978       &  50.4        &  18.9       &    0.776      \\
PredRNN~\cite{wang2017predrnn}       &   56.8     &     126.1      &      0.867        &    46.4      &    17.1      &    0.971      &   48.4       &    18.9     &     0.781     \\
Causal LSTM~\cite{wang2018predrnn++} &   46.5      &    106.8       &    0.898        &     44.8     &     16.9     &     0.977     &    45.8      &     17.2    &      0.851    \\
MIM~\cite{wang2019memory}            &   44.2    &     101.1      &       0.910        &        42.9  &    16.6      &    0.971      &   42.9       &    17.8     &    0.790      \\
E3D-LSTM~\cite{wang2018eidetic}      &   41.3    &      86.4     &       0.920        &     43.2     &      16.9    &     0.979     &    46.4      &     16.6    &     0.869     \\
PhyDNet~\cite{guen2020disentangling} &   24.4      &   70.3        &   0.947      &        41.9  &       16.2   &         0.982 &        36.9  &       16.2  &        0.901  \\
SimVP~\cite{gao2022simvp}            &   23.8      &      68.9     &     0.948             & 41.4         & 16.2         &0.982          & 31.6         & 15.1        &   0.904       \\
MIMO-VP*~\cite{ning2023mimo}        &   17.7      &       51.6    &     0.964             & -        & -         & -          & -         & -        &   -       \\ \hline
IAM4VP      &    \textbf{15.3}       &     \textbf{49.2}      &    \textbf{0.966}       &    \textbf{37.2}     &     \textbf{16.4}     &      \textbf{0.983}           & \textbf{12.6}     &       \textbf{11.2}   &   \textbf{0.942}       \\ \hline \hline
\end{tabular}
}

\end{center}
\caption{Performance comparison results of IAM4VP and recent leading approaches on three future frame prediction common benchmark datasets. IAM4VP achieved state-of-the-art performance on all three benchmark datasets, which have different characteristics: Moving MNIST, TrafficBJ, and Human 3.6. The asterisk (*) indicates performance reported in the author's paper. See Supplement Section 4 for qualitative comparison results.}
\label{tab:common}
\end{table*}
\subsection{Learned Prior (\textit{LP})}
Although random masking of the future queue increases the model's robustness to incomplete future queues, feature vectors extracted from predicted frames still do not reflect feature vectors from actual video frames.
Therefore, the prediction errors may worsen as time step increases in the testing phase.
To address this, we introduce an \textit{LP} that aims to project feature vectors from the predicted results closer to those from actual video frames.

As shown in ~\figref{fig:divmodnet}, \textit{LP} receives the output $\hat{y}_{t}$ of the model $F(\cdot)$ and generates features.
After that, it is trained to minimize the MSE loss between the feature generated through $e(y_{y})$ and the feature generated through the \textit{LP}.
As a result, the \textit{LP} generates a feature that is more similar to the ground truth feature, even in the presence of errors in the predicted results. 
~\figureref{fig:infer} illustrates IAM4VP at inference time.
Note that \textit{LP} is trained through fine-tuning.

\section{Experiments}
In this section, we present the evaluation of IAM4VP on widely used benchmarks for future frame prediction and weather$\slash$climate forecasting, respectively. Additionally, we conducted an ablation study to gain insights into the design of video frame prediction models, and show the qualitative results of video frame interpolation - one of IAM4VP's strengths - in the supplementary materials. Note that IAM4VP also benefits from the implicit model, such as dense inference for future frame prediction.

\paragraph{Training \& Test Datasets}
We evaluate IAM4VP using five datasets, as summarized in ~\tabref{tab:dataset}. Moving MNIST~\cite{srivastava2015unsupervised}, TrafficBJ~\cite{zhang2017deep}, and Human 3.6~\cite{ionescu2013human3} are commonly used benchmark datasets for future frame prediction. Moving MNIST consists of synthetically generated video sequences featuring two digits moving between 0 and 9, while TrafficBJ is a collection of taxicab GPS data and meteorological data recorded in Beijing. Human 3.6 contains motion capture data of a person captured using a high-speed 3D camera. The SEVIR and ICAR-ENSO datasets are weather and climate prediction benchmarks. The SEVIR dataset \cite{veillette2020sevir} comprises radar-derived measurements of vertically integrated liquid water (VIL) captured at 5-minute intervals with 1 km spatial resolution, serving as a benchmark for rain and hail detection. The ICAR-ENSO dataset \cite{ham2019deep} combines observational and simulation data to provide forecasts of El Niño/Southern Oscillation (ENSO), an anomaly in sea surface temperature (SST) in the Equatorial Pacific that serves as a significant predictor of seasonal climate worldwide.

\begin{table}[t!]
\begin{center}
\resizebox{0.9\columnwidth}{!}{%
\begin{tabular}{llllll}
\hline \hline
\textbf{Dataset}      & $N_{Train}$  & ${N_{Test}}$    & (C, H, W)     & $T$  & $\hat{{T}}$ \\ \hline
Moving MNIST & 10,000 & 10,000 & (1,64,64))    & 10 & 10 \\
TrafficBJ    & 19,627 & 1,334  & (2, 32, 32)   & 4  & 4  \\
Human 3.6    & 2,624  & 1,135  & (3, 128, 128) & 4  & 4  \\ \hline
SEVIR        & 35,718 & 12,159 & (1, 384, 384) & 13 & 12 \\
ICAR-ENSO    & 5,205  & 1,667  & (1,24,48)     & 12 & 14 \\ \hline \hline
\end{tabular}%
}
\end{center}
\caption{Statistics of three common benchmark datasets and two weather$\slash$climate forecasting benchmark datasets.}
\label{tab:dataset}
\end{table}
\paragraph{Evaluation metric}
For the evaluation of common benchmark datasets, we adopt widely used evaluation metrics, including Mean Square Error (MSE), Mean Absolute Error (MAE), Peak Signal to Noise Ratio (PSNR), and Structural Similarity Index Measure (SSIM). For rain forecasting models, we use the Critical Success Index (CSI) as an evaluation metric \cite{shi2015convolutional}. In addition, we validate ENSO forecasting using the Nino SST indices \cite{gao2022earthformer}. Specifically, the Nino3.4 index represents the averaged SST anomalies across a specific Pacific region (170$^{\circ}$-120$^{\circ}$W, 5$^{\circ}$S-5N$^{\circ}$N), and defines El Niño/La Niña events based on the SST anomalies around the equator.

\paragraph{Implementation details}
We use the Adam optimizer \cite{kingma2014adam} with $\beta_1=0.9$ and $\beta_2 = 0.999$, and a cosine scheduler without warm-up \cite{loshchilov2016sgdr} in all experiments.
The learning rate is set to 0.001 and the mini-batch size is 16.
To mitigate the learning instability of the model, we apply an exponential moving average (EMA)~\cite{rombach2022high}, which is not typically used in existing future frame prediction models.
The EMA model is updated every 10 iterations, and EMA model updates begin at 2000 iterations.
The EMA model update momentum is set to 0.995.
The Moving MNIST dataset is trained for 10K epochs, and all other datasets are trained for 2K epochs.

\begin{table*}[t!]
\begin{center}
\resizebox{1.9\columnwidth}{!}{%
\begin{tabular}{c|cccccc|ccccc}
\hline \hline
\multirow{2}{*}{\textbf{Model}} & \multicolumn{6}{c|}{\textbf{SEVIR}}                                 & \multicolumn{5}{c}{\textbf{ICAR-ENSO}}                              \\ \cline{2-12} 
                       & \#Param. (M) & GFLOPS & CSI-M  & CSI-160 & CSI-16 & MSE    & \#Param. (M) & GFLOPS & C-Niño3.4-M & C-Niño3.4-WM & MSE   \\ \hline 
UNet                   & 16.6         & 33     & 0.3593 & 0.1278  & 0.6047 & 4.1119 & 12.1         & 0.4    & 0.6926      & 2.102        & 2.868 \\
ConvLSTM               & 14.0         & 527   & 0.4185 & 0.2157  & 0.7441 & 3.7532 & 14.0         & 11.1   & 0.6955      & 2.107        & 2.657 \\
PredRNN                & 23.8         & 328   & 0.4080 & 0.2928  & 0.7569 & 3.9014 & 23.8         & 85.8   & 0.6492      & 1.910        & 3.044 \\
PhyDNet                & 3.1          & 701    & 0.3940 & 0.2767  & 0.7507 & 4.8165 & 3.1          & 5.7    & 0.6646      & 1.965        & 2.708 \\
E3D-LSTM               & 12.9         & 523   & 0.4038 & 0.2708  & 0.7059 & 4.1702 & 12.9         & 99.8   & 0.7040      & 2.125        & 3.095 \\
Rainformer             & 19.2         & 170    & 0.3661 & 0.2675  & 0.7573 & 4.0272 & 19.2         & 1.3    & 0.7106      & 2.153        & 3.043 \\
Earthformer            & 7.6          & 257   & 0.4419 & 0.3232  & 0.7513 & 3.6957 & 7.6          & 23.9   & 0.7329      & 2.259        & 2.546 \\ \hline
IAM4VP          &           34.7   &   392     &   \textbf{0.4607}     &     \textbf{0.3430}    &      \textbf{0.7761}  &       \textbf{2.9371} &     34.2         &   11.8     &   \textbf{ 0.7698}         &       \textbf{2.484}       &  \textbf{1.563}     \\ \hline \hline
\end{tabular}%
}
\end{center}
\label{tab:weather}
\caption{Performance comparison experiment results between IAM4VP and recent leading weather prediction models on SEVIR dataset and ICRA-ENSO dataset. Note that many-to-many models, such as SimVP, could not be tested on the weather prediction benchmark dataset because the model structure must be significantly changed when the input $T$ and output $\hat{T}$ are different.}
\end{table*}

\subsection{Common Benchmark Results}
\paragraph{Moving MNIST}
The first row of ~\tabref{tab:common} shows the performance comparison results of IAM4VP with recent leading approaches on the Moving MNIST dataset.
PhyDNet and SimVP are models that achieve state-of-the-art results in autoregressive and non-autoregressive methods, respectively.
As shown in ~\tabref{tab:common}, IAM4VP achieves MSE \textbf{15.3}, MAE \textbf{51.6}, and SSIM \textbf{0.965} on the Moving MNIST dataset, outperforming both autoregressive and non-autoregressive methods on this rule-based synthetic dataset.

\paragraph{TrafficBJ}
The second row of ~\tabref{tab:common} presents the performance comparison results of IAM4VP and other approaches on the TrafficBJ dataset.
In most cases of the TrafficBJ dataset, the past and future have a linear relationship, resulting in saturated values for the MSE, MAE, and SSIM of all approaches.
Nonetheless, IAM4VP achieves state-of-the-art results on the TrafficBJ dataset by a large margin.
These experimental results indicate that IAM4VP is also effective on real-world datasets with relatively easy linear relationships.

\paragraph{Human 3.6}
The third row of ~\tabref{tab:common} displays the performance of IAM4VP on the Human 3.6 dataset.
Human 3.6 has a non-linear relationship between the past and future because it has to predict the future behavior of humans.
As shown in the table, IAM4VP outperforms previous models, with a performance improvement of 19.0 in the MSE metric.
These experimental results demonstrate that IAM4VP is particularly robust on real-world datasets that require non-linear modeling.

\subsection{Weather$\slash$Climate Benchmark Results}
\paragraph{SEVIR}
SEVIR data has a higher resolution than MovingMNIST, and the shape of water objects is less distinct than that of numbers and humans.
Although Earthformer had already achieved state-of-the-art on this dataset, our model achieved better scores, with a CSI-M of \textbf{0.4607} and MSE of \textbf{2.9371} compared to Earthformer, as shown in Table~\ref{tab:weather}. The CSI index is generally used to validate the accuracy of the model precipitation forecast and is defined as CSI = Hits / (Hits+Misses+F.Alarms). The CSI-160 and CSI-16 calculate Hits(obs = 1, pred = 1), Misses(obs=1, pred=0), and False Alarms(obs = 0, pred = 1) based on binary thresholds from 0-255 pixel values. We used thresholds [16, 74, 133, 160, 181, 219], and the CSI-M is their averaged value \cite{gao2022earthformer}.

\begin{table}[t!]
\begin{center}
\resizebox{\columnwidth}{!}{%
\begin{tabular}{l|c|c|c}
\hline \hline
\multicolumn{1}{c|}{\textbf{Component}} & \textbf{\#Param .(M)} & \textbf{MSE} & \textbf{Training Epoch} \\ \hline
SimVP                          &    20.4     & 23.5    & 2K             \\
+Improved Autoencoder          &    20.4     &  23.0   & 2K             \\
+ConvNeXt (ST)                 &    20.4     &  18.2   & 2K             \\ \hline
+STR                           &    20.5     &   17.6  & 2K             \\ 
+STR                           &    20.5     &   \textbf{16.2}  & 10K             \\ \hline
+Time Step MLP Embedding       &   20.6      &   19.3  & 2K            \\ 
+Time Step MLP Embedding       &   20.6      &   \textbf{17.4}  & 10K            \\ \hline
+Stacked Autoregressive  w/o $M_{g}$    &    20.7     &   25.2  & 10K          \\
+Stacked Autoregressive  w/  $M_{g}$   &    20.7     &   16.9  & 2K            \\
+Stacked Autoregressive  w/  $M_{g}$   &    20.7     &   \textbf{15.8}  & 10K           \\ \hline
+Learned Prior (IM4VP)                       &  20.7       &   15.9  & 2K+Fine-tune \\
+Learned Prior (IM4VP)                       &  20.7       &   \textbf{15.3}  & 10K+Fine-tune 
\\ \hline \hline
\end{tabular}%
}
\end{center}
\caption{Performance analysis for IAM4VP components.}
\label{tab:able1}
\end{table}

\paragraph{ICAR-ENSO}
The forecast evaluation, as C-Nino3.4, is calculated using the correlation skill (\cite{ham2019deep}) of the three-month-averaged Nino3.4 index. We forecast up to 14-month SST anomalies (2 months more than input data for calculating three-month-averaging) from the 12-month SST anomaly observation. As shown in Table~\ref{tab:weather}, IAM4VP outperforms all other methods on all metrics. The C-Nino3.4-M is the mean of C-Nino3.4 over 12 forecasting steps, and C-Nino3.4-WM is the time-weighted mean correlation skill. We also computed MSE to evaluate the spatio-temporal accuracy between prediction and observation. Our model also improved the C-Nino3.4 indexes, but what is noteworthy is that the MSE is reduced by over 1 compared to other models.

\begin{figure}[t!]
    \centering
    \includegraphics[width=1.0\columnwidth]{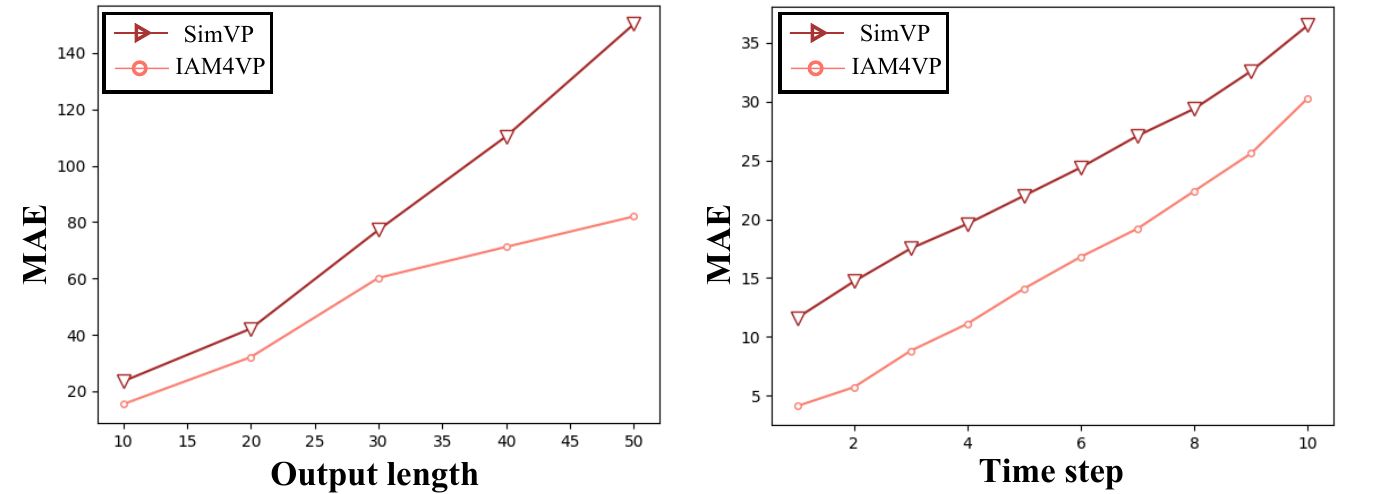}
    \caption{Performance comparison experiment according to time step and output length changes of SimVP and IAM4VP on the Moving MNIST dataset.}
    \label{fig:output}
\end{figure}

\subsection{Ablation Study}
\paragraph{Component effect}
We conducted an ablation study on the Moving MNIST dataset to analyze the effect of each component of IAM4VP. The results of this study are presented in Table 1, where we analyzed the performance of IAM4VP by sequentially adding each component to the SimVP baseline.

In SimVP, we observed a performance improvement of 0.5 by changing the encoder structure from {\fontfamily{qcr}\selectfont(Conv, GroupNorm, LeakyReLU)} to {\fontfamily{qcr}\selectfont(Conv, LayerNorm, SiRU)} and the decoder structure from {\fontfamily{qcr}\selectfont(ConvTranspose, GroupNorm, LeakyReLU)} to {\fontfamily{qcr}\selectfont(Conv, LayerNorm, SiRU, PixelShuffle)}.
Furthermore, by changing the spatio-temporal predictor from the existing inception block to the ConvNeXt block, we observed a significant performance improvement of +4.8.
These experimental results indicate the importance of the spatio-temporal predictor in the video frame prediction task.

Next, we applied STR~\cite{seo2022simple}, a refinement module that demonstrated good performance on weather forecast datasets, resulting in a performance improvement of 0.6. By changing the model from the MIMO to the implicit MISO and extending the training schedule from 2K to 10K epochs, we achieved an additional performance boost of +0.2. While this improvement may seem small, it was crucial to implementing the stacked autoregressive model structure.

\paragraph{Output condition dependency}
The experimental results of our analysis on the performance according to the output condition are presented in ~\figureref{fig:output}.
The left graph shows a performance comparison between SimVP and IAM4VP according to the output length.
As shown in the figure, SimVP exhibits a linearly increasing error as the output length increases, whereas IAM4VP does not, indicating that IAM4VP is more robust to the output length.

The right graph shows the performance analysis result according to time steps, where the error in both methods increases as the time step increases, as expected.
These experimental results demonstrate that predicting long output lengths is more challenging, which increases the difficulty of the prediction.
Overall, the experimental results suggest that IAM4VP outperforms SimVP when predicting long output lengths.
Specifically, IAM4VP's error increases linearly up to output length 20 (i.e., time step 20), but does not increase linearly beyond this point.
These findings underscore the importance of the IAM4VP's design, and the components of this model can be usefully applied in tasks such as weather forecasting.

\subsection{Qualitative Results}
Due to page limitations, please refer to the supplementary material for a more detailed qualitative analysis of the Moving MNIST and ICRA-ENSO datasets.

\begin{figure}[t!]
    \centering
    \includegraphics[width=1.0\columnwidth]{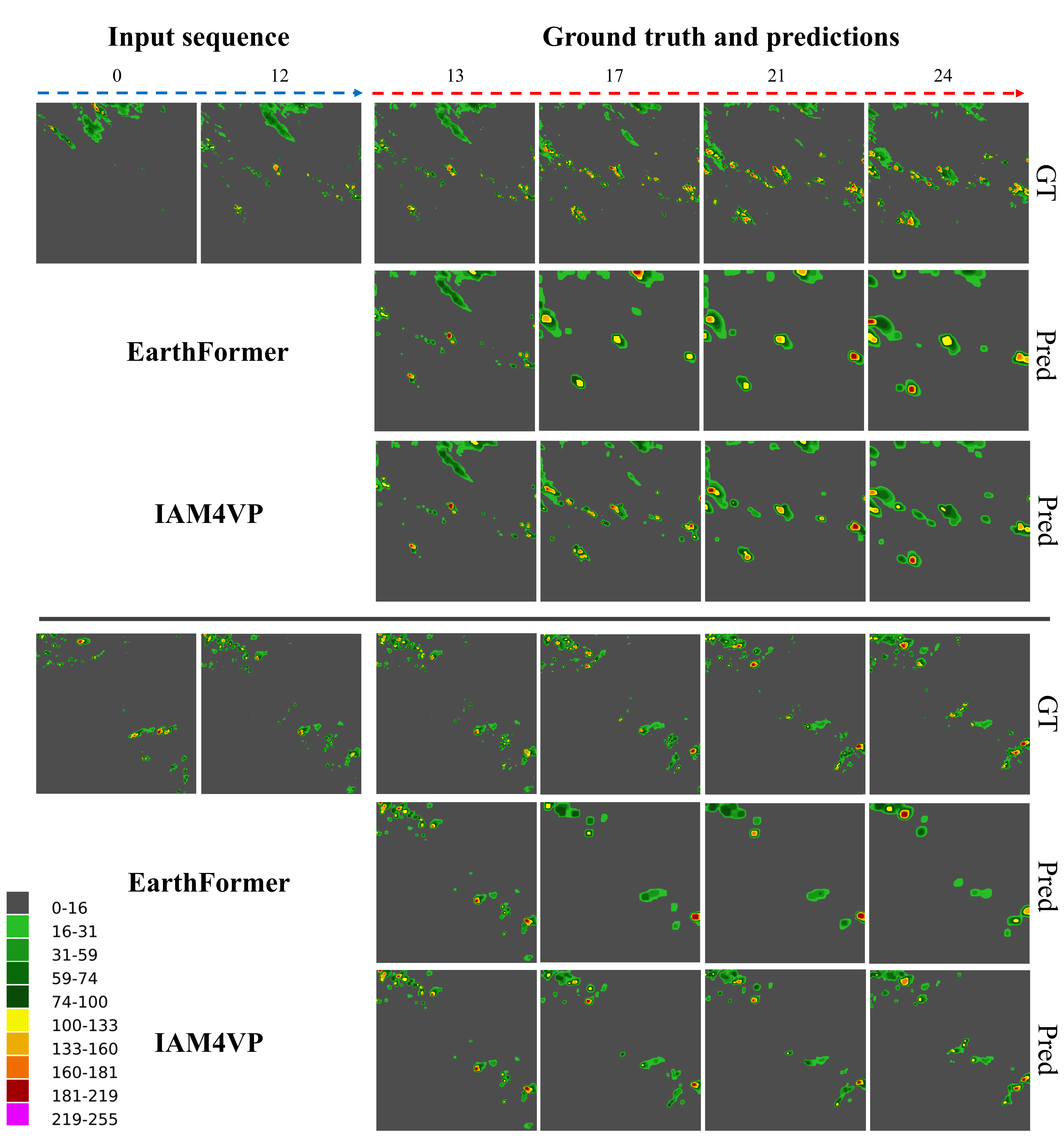}
    \caption{Prediction results of IAM4VP and Earthformer on the SEVIR dataset, represented by vertically integrated liquid water contents (0-255 scale) shown on the color bar.}
    \label{fig:seivr}
\end{figure}

\paragraph{SEVIR}
~\figref{fig:seivr} presents the predicted results of IAM4VP and Earthformer on the SEVIR dataset.
This dataset represents radar-estimated liquid water in a vertical air column, and the color bar indicates water mass per unit area.
A high VIL value corresponds to high water content and probability of precipitation or even the presence of hail in storms.
We predict 12 future frames (four samples on the right side) from 12 input images (two on the left).
In both examples, both models predict plausible results in the first future frame.
However, there is a difference in the results from frame 17 onwards.
Earthformer shows blurry issues, which are simplified boundaries of cells and removes small-sized cells.
This is known as a general limitation of deep-learning-based weather prediction models.
They underestimate VIL values more than IAM4VP but predict broader regions with high water content.
On the other hand, IAM4VP accurately predicts cell edges up to frame 24.
In particular, even small-sized cells are conserved during training.
This means that some cells not visible in Earthformer's results can be predicted in our model.

\begin{figure*}[hbt!]
    \centering
    \includegraphics[width=2.0\columnwidth]{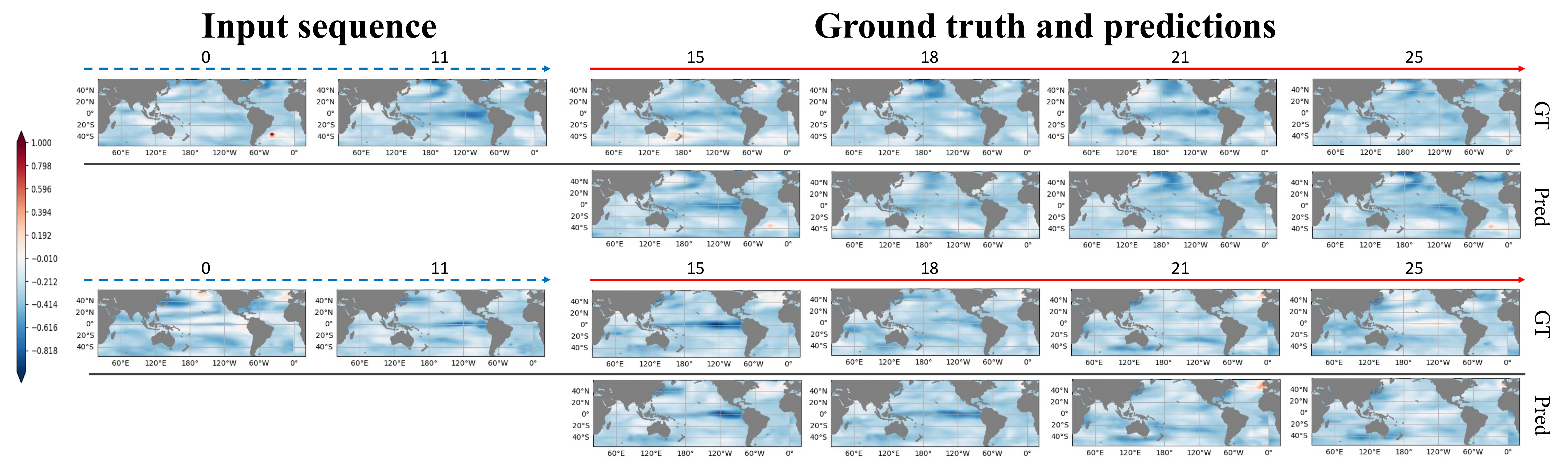}
    \caption{Prediction results of IAM4VP on the ICRA-ENSO dataset. The color bar means SST anomalies on the global map.}
    \label{fig:ffi}
\end{figure*}
\section{Conclusion}
In this paper, we propose a novel implicit video prediction model, called IAM4VP, based on the new autoregressive method called stacked autoregressive. The stacked autoregressive method is designed to solve the error accumulation problem that occurs as the time step increases in existing autoregressive methods. Unlike existing methods, the observed frame information is not lost and is maintained until the end, even if the time step increases. We achieved state-of-the-art results on five public benchmark datasets for video prediction tasks using IAM4VP, which is designed based on the MISO-Multi Model, which has the highest performance in video prediction tasks. Furthermore, we find that our model can be extended to the field of deep learning-based weather and climate forecasting with additional advantages such as temporal interpolation. 

\appendix

\section{Appendix}

\subsection{Implementation details}
IAM4VP consists of an encoder $e(.)$, a spatial-temporal predictor $p(.)$, a spatial decoder $d(.)$, a spatial-temporal refinement module (STR), and a positional encoder $s(.)$.
We have empirically confirmed that the amount of computation (Flops) and accuracy depending on the number of $e(.)$, $d(.)$, and $p(.)$.
~\tableref{tab:dd} shows the number of channels and layers for each dataset.
Most of the hyperparameters were set according to SimVP's~\cite{gao2022simvp} recipe. We empirically found that for larger image sizes, it is more efficient to use more encoders and decoders, while for smaller image sizes, it is more efficient to reduce the number of encoders and decoders.
Code is available at https://github.com/seominseok0429/Implicit-Stacked-Autoregressive-Model-for-Video-Prediction.

\begin{table}[h!]
\begin{center}
\resizebox{\columnwidth}{!}{%
\begin{tabular}{c|ccccc}
\hline \hline
                   & \textbf{Moving MNIST} & \textbf{TrafficBJ} & \textbf{Human 3.6} & \textbf{SEVIR} & \textbf{ICRA-ENSO} \\ \hline
$e(.)_{channels}$ &        64      &   64        &    64       &   64    &   64        \\
$d(.)_{channels}$ &        64      &     64      &     64      &    64   &     64      \\
$p(.)_{channels}$ &        512      &    256       &   128        &   512    &  128         \\
$e(.)_{num}$      &     4         &   3        &    4       &  5     &    2      \\
$d(.)_{num}$      &     4        &    3       &    4       &   5    &    2       \\
$p(.)_{num}$      &  6            &     4      &    8       &   6    &     4      \\ \hline \hline
\end{tabular}
}
\end{center}
\caption{IAM4VP hyperparameter settings of each component.}
\label{tab:dd}
\end{table}
\subsection{Qualitative Results in ICRA-ENSO dataset}
We predict the ICAR-ENSO dataset in  ~\figref{fig:ffi}, and the results show the global map of SST anomalies.
In the color bar, the blue side means negative SST anomalies, and the red side means positive.
%
%
These results show that IAM4VP slightly overestimates values but predicts anomalous patterns that change from negative to positive according to the time, similar to observations (GT).
%
%
%
In the early stages of prediction, even if the model predicts incorrect results in some regions, subsequent predictions show the correct SST anomaly pattern again.  
This means the IAM4VP model does not propagate early prediction errors to later predictions. 

\subsection{Qualitative Comparison Results}\label{sec:qual}
\begin{figure*}[t!]
    \centering
    \includegraphics[width=1.7\columnwidth]{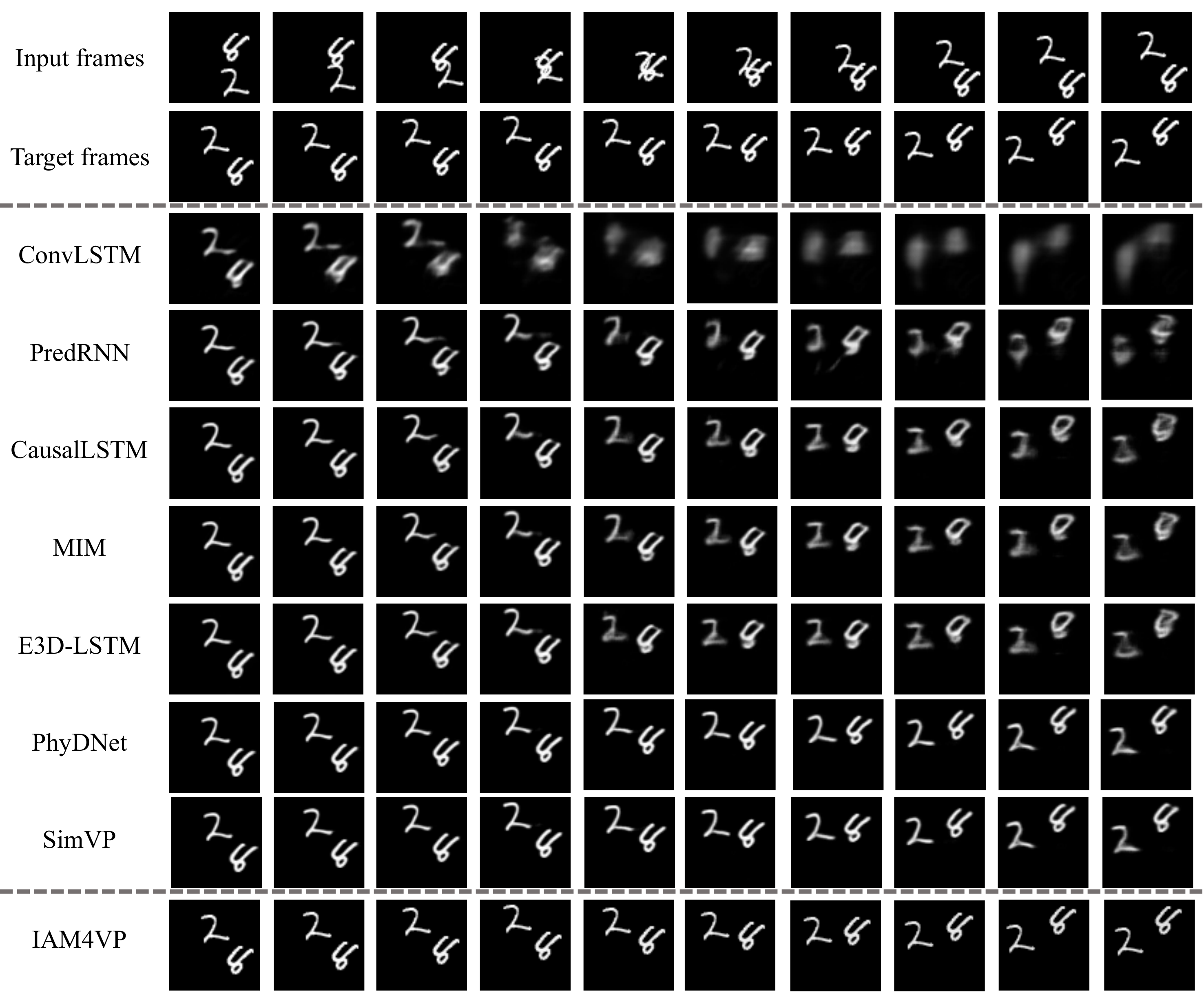}
    \caption{The qualitative comparison experiments between commonly used video frame prediction models and IAM4VP on the Moving MNIST dataset.}
    \label{fig:ab2}
\end{figure*}

~\figureref{fig:ab2} presents a qualitative comparison between existing video frame prediction methods and IAM4VP.
As shown in the ~\figref{fig:ab2}, autoregressive models~\cite{xu2018predcnn, wang2018predrnn++, shi2015convolutional} commonly used in video frame prediction suffer from accumulated errors and blurry images as the lead time increases.
In contrast, SimVP and IAM4VP produce relatively clear images across all lead times, with IAM4VP in particular yielding the clearest images.

\subsection{Does IAM4VP really consider the correlation between each lead time output?}
\begin{figure*}[hbt!]
    \centering
    \includegraphics[width=1.6\columnwidth]{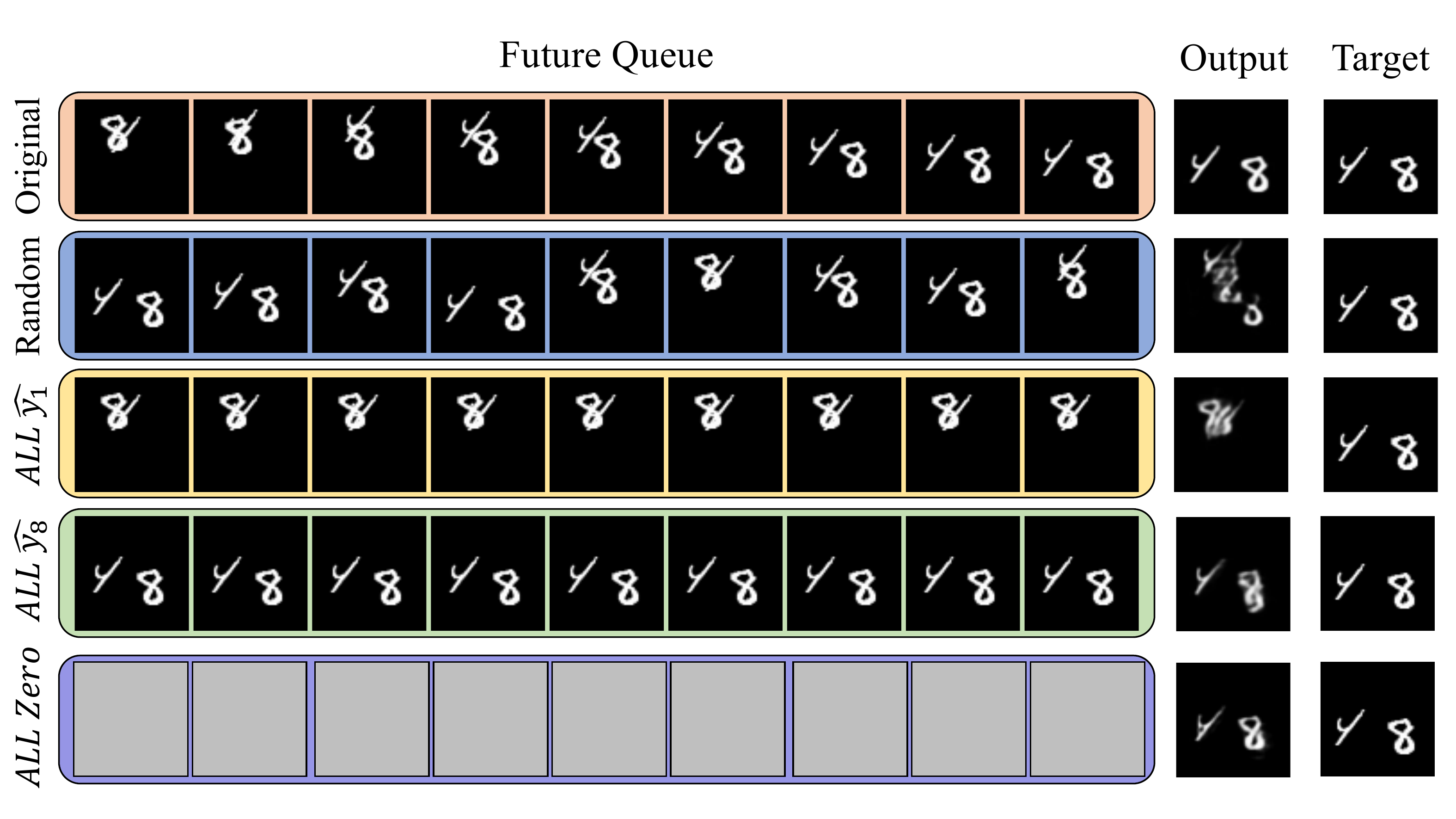}
    \caption{Qualitative comparison results of IAM4VP output according to future queue configuration. \textit{Original} represents stacking the future queue in the correct order, \textit{Random} represents random shuffling of the future queue, \textit{All} $\hat{y_{1}}$ and \textit{All} $\hat{y_{8}}$ indicate using only $\hat{y_{1}}$ and $\hat{y_{8}}$, respectively, for the entire future queue, and \textit{All zero} represents not using the future queue at all. Note that the value of $t$ is set to 9.}
    \label{fig:a1}
\end{figure*}

IAM4VP uses future queues for inference, but does not depend on future queues.
IAM4VP utilizes future queues for inference, but it does not depend solely on them.
Therefore, one might naturally question whether IAM4VP relies only on observed queue information for predictions.
To address this question, we conducted qualitative experiments.
~\figureref{fig:a1} illustrates the outputs of IAM4VP for different future queue configurations.
As seen in the results of the \textit{Random} experiment in ~\figref{fig:a1}, randomly shuffling the future queue leads to inaccurate IAM4VP outputs.
This experiment result indicates that IAM4VP is affected by the future queue.
Additionally, the best performance was achieved when the future queue was stacked in the correct order (\textit{Original}).
However, as demonstrated in the \textit{All Zero} experiment, IAM4VP is still capable of generating plausible images even when the future queue is not used.

\begin{figure*}[t!]
    \centering
\includegraphics[width=1.6\columnwidth]{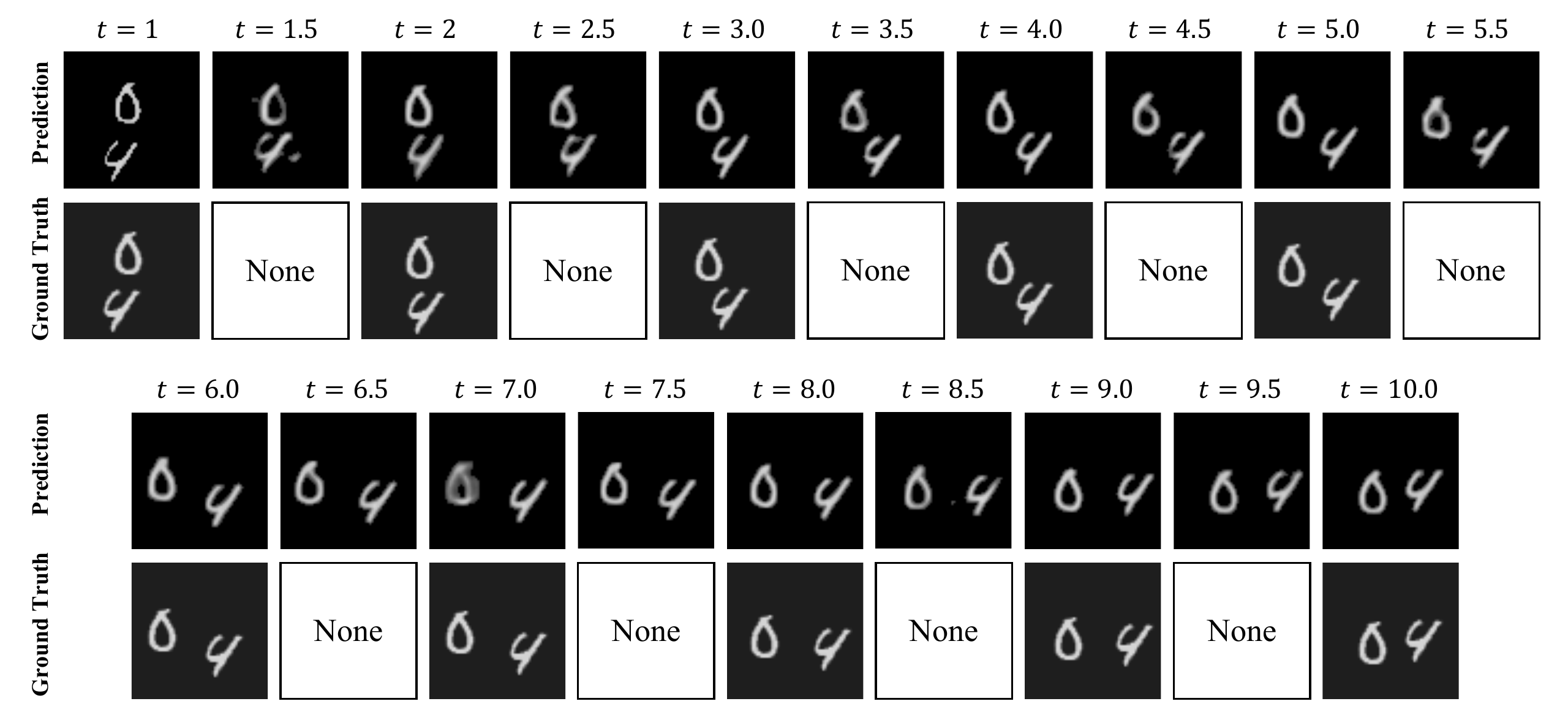}
    \caption{Future frame prediction and interpolation quantitative experiment results on the Moving MNIST dataset. Since IAM4VP is an implicit model, it was trained with a lead time interval of 1, but it can operate at an interval of 0.5 during inference.}
    \label{fig:ab3}
\end{figure*}
\begin{figure*}[t!]
    \centering
\includegraphics[width=1.6\columnwidth]{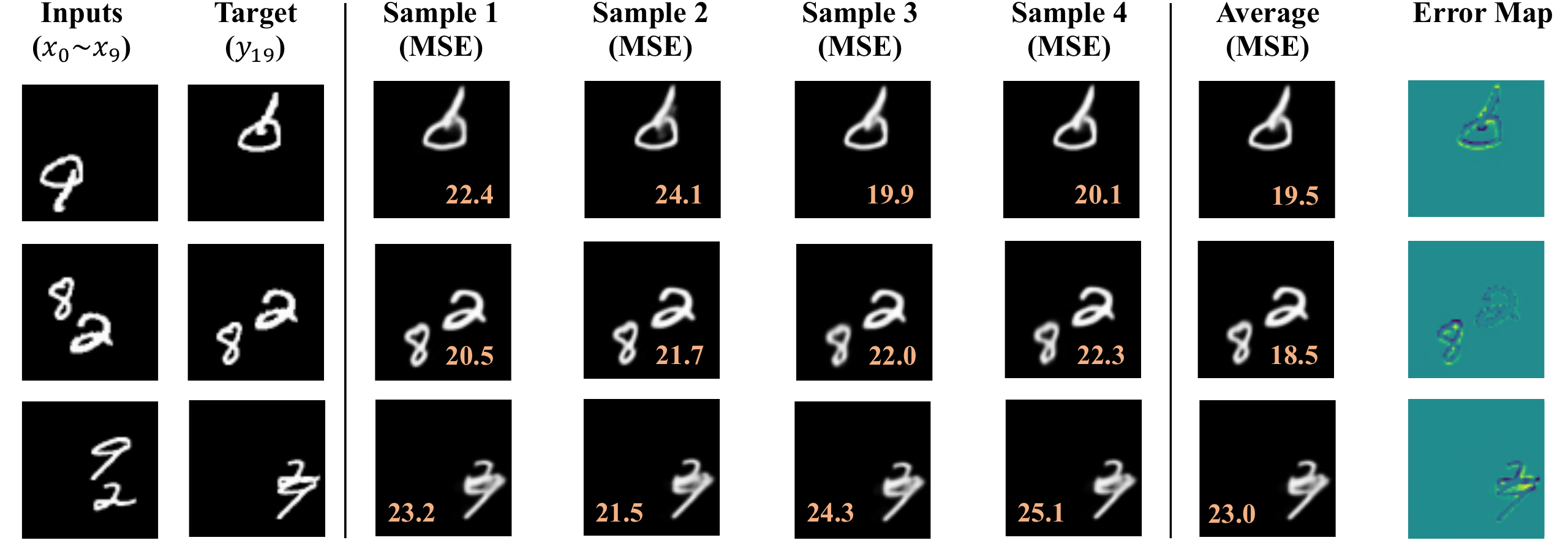}
    \caption{IAM4VP's diverse prediction. Note that Average is an average of the four samples shown in the figure, not an ensemble of all possible samples.}
    \label{fig:ab4}
\end{figure*}
\subsection{Dense Inference for VP}
IAM4VP is an implicit model for time variables.
Therefore, IAM4VP, like implicit neural representations models, can make dense predictions regardless of the interval $t$ of the training dataset.
~\figureref{fig:ab3} shows the qualitative results of inference at intervals of $\frac{t}{2}$ in the Moving MNIST Dataset.
As shown in the figure, it was shown that IAM4VP can perform video temporal interpolation differently from existing video frame prediction models~\cite{wang2017predrnn, wang2018predrnn++, castrejon2019improved, gao2022simvp}.
The advantages of implicit model are especially needed for weather prediction tasks.
The temporal resolution of most geostationary-satellites is not good, and a lot of cost is consumed to increase the resolution.
Past satellites have low temporal resolution, while current satellites have higher temporal resolution.
Therefore, a model capable of training on irregular time steps is required to learn from observations from the past to the present.
For this reason, IAM4VP holds potential for application in geostationary satellites.
In our future work, we plan to explore the application of IAM4VP in geostationary satellites.
\subsection{Diverse Future Prediction}
IAM4VP can generate various output combinations from a single model depending on the usage of each index in the future queue.
IAM4VP can generate various output combinations from a single model depending on the usage of each index in the future queue.
~\figureref{fig:ab4} shows four different samples produced based on different future queue combinations when the target $t$ is 9. (Note that when $t=9$, there are a total of 8 possible indices in the future queue, resulting in $2^{8}$ possible combinations.)
As shown in the ~\figref{fig:ab4}, IAM4VP is capable of generating samples with distinct levels of MSE, which can then be ensembled.
While performance significantly improves in some cases (as seen in the second column of the figure), there are also cases where ensemble performance does not improve significantly (as seen in the third column of the figure).
In our future work, we plan to conduct research on finding optimal ensemble policies.
\section{Limitation and Future Work}
\paragraph{Output length flexibility}
Unlike the existing autoregressive model, the stacked autoregressive method lacks flexibility for lead time.
For example, the lead time of the existing autoregressive model can be increased as much as desired by adjusting the autoregressive step even if the performance is degraded.
However, since the stacked autoregressive method has to fix the length of the feature map input to the model, the length of the lead time cannot exceed the existing fixed length.
However, the stacked autoregressive method can also perform inference in the same way as the existing autoregressive methods, but this does not match the motivation of the stacked autoregressive method.
In our future work, we will conduct research on increasing the output length flexibility of the stacked autoregressive method.

\paragraph{Long training time}
Since IAM4VP is an implicit model, it requires a longer training time than existing MIMO models.
When the size of the dataset is small or the resolution is small, such as Moving MNIST, TrafficBJ, Human 3.6, and ICAR-ENSO, the increased training time is not burdensome, but in the case of the SEVIR dataset, about 18 days of additional training time was required.
In our future work, we plan to conduct research on reducing the convergence time of IAM4VP.%

{\small
\bibliographystyle{ieee_fullname}
\bibliography{egbib}
}

\end{document}